\documentclass{article} 
\usepackage{iclr2019_conference,times}

\usepackage{amsmath,amsfonts,bm}

\def\eqref#1{equation~\ref{#1}}

\def\1{\bm{1}}

\DeclareMathAlphabet{\mathsfit}{\encodingdefault}{\sfdefault}{m}{sl}
\SetMathAlphabet{\mathsfit}{bold}{\encodingdefault}{\sfdefault}{bx}{n}

\usepackage{hyperref}
\usepackage{url}
\usepackage{amssymb,amsmath, breqn, bm, amsthm}
\usepackage{listings}
\usepackage{color}
\usepackage{courier}
 
\definecolor{mygreen}{rgb}{0,0.6,0}
\definecolor{mygray}{rgb}{0.5,0.5,0.5}
\definecolor{mymauve}{rgb}{0.58,0,0.82}

\usepackage{mathtools}
\usepackage{appendix}
\usepackage{wrapfig}
\usepackage[ruled,vlined]{algorithm2e}
\usepackage{bbm}
\newtheorem{theorem}{Theorem}[section]

\newtheorem{corollary}{Corollary}[theorem]
\newtheorem{lemma}[theorem]{Lemma}
\theoremstyle{definition}
\newtheorem{definition}{Definition}[section]
\DeclareMathOperator{\clip}{clip}
\usepackage {graphicx, epigraph}
\usepackage{comment}
\usepackage{subcaption}
\usepackage{booktabs}

\usepackage{color}

\def\viewcomments{0}

\ifx\viewcomments\undefined
  \newcommand{\AT}[1]{\iffalse{[Aviv: #1]}\fi}
  \newcommand{\ES}[1]{\iffalse{[Elad: #1]}\fi}
  \newcommand{\SK}[1]{\iffalse{[Sarit: #1]}\fi}
  \newcommand{\SA}[1]{\iffalse{[Shira: #1]}\fi}
\else
  \newcommand{\AT}[1]{\textbf{\textcolor{green}{[Aviv: #1]}}}
  \newcommand{\ES}[1]{\textbf{\textcolor{blue}{[Elad: #1]}}}
  \newcommand{\SK}[1]{\textbf{\textcolor{red}{[Sarit: #1]}}}
  \newcommand{\SA}[1]{\textbf{\textcolor{magenta}{[Shira: #1]}}}
\fi

\title{Constrained Policy Improvement for Safe and Efficient Reinforcement Learning}

\author{Elad Sarafian \\
Bar-Ilan University, Israel \\
\texttt{elad.sarafian@gmail.com}
\And
Aviv Tamar \\
UC Berkeley \\
\texttt{avivt@berkeley.edu} \\
\AND
Sarit Kraus \\
Bar-Ilan University, Israel \\
\texttt{sarit@cs.biu.ac.il}
}

\iclrfinalcopy 
\begin{document}

\maketitle

\begin{abstract}

      We propose a policy improvement algorithm for Reinforcement Learning (RL) which is called Rerouted Behavior Improvement (RBI). RBI is designed to take into account the evaluation errors of the Q-function. Such errors are common in RL when learning the $Q$-value from finite past experience data. Greedy policies or even constrained policy optimization algorithms which ignore these errors may suffer from an improvement penalty (i.e. a negative policy improvement). To minimize the improvement penalty, the RBI idea is to attenuate rapid policy changes of low probability actions which were less frequently sampled. This approach is shown to avoid catastrophic performance degradation and reduce regret when learning from a batch of past experience. Through a two-armed bandit with Gaussian distributed rewards example, we show that it also increases data efficiency when the optimal action has a high variance. We evaluate RBI in two tasks in the Atari Learning Environment: (1) learning from observations of multiple behavior policies and (2) iterative RL. Our results demonstrate the advantage of RBI over greedy policies and other constrained policy optimization algorithms as a safe learning approach and as a general data efficient learning algorithm.  An anonymous Github repository of our RBI implementation is found at \url{https://github.com/eladsar/rbi}.

\end{abstract}

\section{Introduction}

While Deep Reinforcement Learning (DRL) is the backbone of many of the recent Artificial Intelligence breakthroughs \citep{silver2017mastering,openai_2018}, it suffers from several factors which inhibit deployment of RL systems to real-world tasks. Two of these elements are: (1) data efficiency and; (2) safety. DRL is notoriously data and time inefficient, requires up to billions of history states \citep{horgan2018distributed} or weeks of wall-clock time \citep{hessel2017rainbow} to train to expert level. While it is partially due to the slow training process of deep neural networks, it is also due to inefficient, yet simple to implement, policy improvement routines. For example, a greedy policy improvement (with a fix exploration parameter) is known to have a higher regret than other methods such as Upper Confidence Bound (UCB) \citep{auer2002finite}, but the latter is much more difficult to adjust to a deep learning framework \citep{bellemare2016unifying}. This transformation from the countable state space of bandit and grid-world problems to the uncountable state-space in a DRL framework, calls for efficient improvement methods which fit into existing deep learning frameworks.

For some real-world problems, like autonomous cars \citep{shalev2016safe}, safety is a crucial factor. Random initialized policies and even a RL algorithm that may suffer from sudden catastrophic performance degradation are both unacceptable in such environments. While policy initialization may be solved with Learning from Demonstrations (LfD) algorithms \citep{argall2009survey}, changing the policy in order to improve performance is still a risky task. Largely since the $Q$-value of a current policy can only be estimated from the past data. Therefore, for safe RL, it is desirable to design improvement algorithms that model the accuracy of the $Q$-value evaluation and can mitigate between fast improvement and a safety level \citep{garcia2015comprehensive,thomas2015high}.

In this work, we propose a policy improvement method that addresses both the sample efficiency of the learning process and the problem of safe learning from incomplete past experience. We start by analyzing the improvement penalty of an arbitrary new policy $\pi(a|s)$ based on an estimated Q-function of a past behavior policy $\beta(a|s)$. We find that under a simplified model of learning the $Q$-values from i.i.d samples, the variance of a potential improvement penalty is proportional to $\frac{|\beta(a|s)-\pi(a|s)|^2}{\beta(a|s)}$. Therefore, we design a constraint, called reroute, that limits this term. We show that finding the optimal policy under the reroute constraint amounts to solving a simple linear program. Instead of optimizing this policy via a gradient descent optimization, we take a different approach and solve it in the non-parameterized space for every new state the actor encounters. In order, to learn the new improved policy with a parameterized Neural Network (NN), we store the calculated policy into a replay buffer and imitate the actor's policy with a KL regression.

While RBI is designed for safe learning from a batch of past experience, we show that it also increase data efficiency with respect to a greedy step and other constraints such as the Total Variation (TV) \citep{kakade2002approximately} and PPO \citep{schulman2017proximal}. In fact it is akin in practice to the forward KL constraint \citep{vuong2018supervised}, however, unlike the KL constraint, it does not require different scaling for different reward signals and it is much more intuitive to design. We validate our findings both in simple environments such as a two-armed bandit problem with Gaussian distributed reward and also in a complex distributed actors framework when learning to play Atari. 

\section{Rerouted Behavior Improvement}
\label{policy_improvement}

Let us start by examining a single improvement step from a batch of past experience of a behavior policy. Define by $\beta$ the behavior policy of a dataset $\mathcal{D}$ and by $Q^{\beta}$, and $\hat{Q}^{\beta}$ its true and approximated Q-functions. Theoretically, for an infinite dataset with infinite number of visitations in each state-action pair, one may calculate the optimal policy in an off-policy fashion \citep{watkins1992q}. However, practically, one should limit its policy improvement step over $\beta$ when learning from a realistic finite dataset. To design a proper constraint, we analyze the statistics of the error of our evaluation of $\hat{Q}^{\beta}$. This leads to an important observation: the $Q$-value has a higher error for actions that were taken less frequently, thus, to avoid improvement penalty, we must restrict the ratio of the change in probability $\frac{\pi}{\beta}$. We will use this observation to craft the reroute constraint, and show that other well-known monotonic improvement methods (e.g. PPO and TRPO) overlooked this consideration, hence they do not guarantee improvement when learning from a finite experience.  

\subsection{Soft Policy Improvement}
Before analyzing the error's statistics, we begin by considering a set of policies which improve $\beta$ if our estimation of $Q^{\beta}$ is exact. Out of this set we will  pick our new policy $\pi$.  Recall that the most naive and also common improvement method is taking a greedy step, i.e. deterministically acting with the highest $Q$-value action in each state. This is known by the policy improvement theorem \citep{Sutton2016ReinforcementL}, to improve the policy performance. The policy improvement theorem may be generalized to include a larger family of soft steps.
\begin{lemma}[Soft Policy Improvement]
\label{my_policy_improvement}
Given a policy $\beta$, with value and advantage $V^{\beta},A^{\beta}$, a policy $\pi$ improves $\beta$,  i.e. $V^{\pi} \geq V^{\beta} \ \forall s$, if it satisfies $\sum_a \pi(a|s) A^{\beta}(s,a) \geq 0 \ \forall s$ with at least one state with strict inequality. The term $\sum_a \pi(a|s) A^{\beta}(s,a)$ is called the improvement step.\footnote{The proof adhere to the same steps of the greedy improvement proof in \citep{Sutton2016ReinforcementL}, thus it is omitted for brevity.}
\end{lemma}
Essentially, every policy that increases the probability of taking positive advantage actions over the probability of taking negative advantage actions achieves improvement. Later, we will use the next Corollary to prove that RBI guarantees a positive improvement step.
\begin{corollary}[Rank-Based Policy Improvement]
\label{rank_base}
Let $(A_i)_{i=1}^{|\mathcal{A}|}$ be an ordered list of the $\beta$ advantages in a state $s$, s.t. $A_{i+1} \geq A_{i}$, and let $c_i=\pi_i/\beta_i$. If for all states $(c_i)_{i=1}^{|\mathcal{A}|}$ is a monotonic non-decreasing sequence s.t. $c_{i+1} \geq c_{i}$, then $\pi$ improves $\beta$ \textit{(Proof in the appendix)}.
\end{corollary}
 
\subsection{Standard Error of the Value Estimation}
\label{value_se}

To provide a statistical argument for the expected error of the $Q$-function, consider learning $\hat{Q}^{\beta}$ with a tabular representation. The $Q$-function is the expected value of the random variable $z^{\pi}(s,a) = \sum_{k \geq 0} \gamma^k r_k | s,a,\pi$. Therefore, the Standard Error (SE) of an approximation $\hat{Q}^{\beta}(s,a)$ for the $Q$-value with $N$ i.i.d. MC trajectories is
\begin{equation}
\sigma_{\varepsilon(s,a)} = \frac{\sigma_{z(s,a)}}{\sqrt{N_s\beta(a|s)}},
\end{equation}
where $N_s$ is the number of visitations in state $s$ in $\mathcal{D}$, s.t. $N=\beta(a|s)N_s$. Therefore, $\sigma_{\varepsilon(s,a)}\propto\frac{1}{\sqrt{\beta(a|s)}}$ and specifically for low frequency actions such estimation may suffer large SE.\footnote{Note that even for deterministic environments, a stochastic policy inevitably provides $\sigma_{z(s,a)} > 0$.} Notice that we ignore recurrent visitations to the same state during the same episode and hence, the MC discounted sum of rewards are independent random variables.

\subsection{Policy Improvement in the Presence of Value Estimation Errors}
We now turn to the crucial question of what happens when one applies an improvement step with respect to an inaccurate estimation of the $Q$-function, i.e. $\hat{Q}^{\beta}$.
\begin{lemma}[Improvement Penalty]
\label{improvement_penalty}
Let $\hat{Q}^{\beta} = \hat{V}^{\beta} + \hat{A}^{\beta}$ be an estimator of $Q^{\beta}$ with an error $\varepsilon(s,a) = (Q^{\beta} - \hat{Q}^{\beta})(s,a)$ and let $\pi$ be a policy that satisfies lemma \ref{my_policy_improvement}  with respect to $\hat{A}^{\beta}$. Then the following holds

\begin{equation}
\label{improvement_penalty_eq}
V^{\pi}(s) - V^{\beta}(s) \geq  -\mathcal{E}(s) = 
- \sum_{s'\in\mathcal{S}}\rho^{\pi}(s'|s)\sum_{a\in\mathcal{A}}\varepsilon(s',a)\left(\beta(a|s')-\pi(a|s')\right),
\end{equation}
where $\mathcal{E}(s)$ is called the improvement penalty and $\rho^{\pi}(s'|s)=\sum_{k \geq 0} \gamma^k P(s\xrightarrow[]{k}s'|\pi))$ is the unnormalized discounted state distribution induced by policy $\pi$ \textit{(proof in the appendix)}.
\end{lemma}

Since $\varepsilon(s',a)$ is a random variable, it is worth to consider the variance of $\mathcal{E}(s)$. Define each element in the sum of Eq. (\ref{improvement_penalty_eq}) as $x(s',a; s) = \rho^{\pi}(s'|s)\varepsilon(s,a)(\beta(a|s')-\pi(a|s'))$. The variance of each element is therefore
\begin{equation*}
    \sigma^2_{x(s',a;s)} = (\rho^{\pi}(s'|s))^2 \sigma^2_{\varepsilon(s',a)} (\beta(a|s')-\pi(a|s'))^2 =  \frac{(\rho^{\pi}(s'|s))^2\sigma^2_{z(s',a)}}{N_{s'}}\frac{(\beta(a|s')-\pi(a|s'))^2}{\beta(a|s')}.
\end{equation*}
To see the the need for the reroute constraint, we can bound the total variance of the improvement penalty
\begin{equation*}
    \sum_{s',a} \sigma^2_{x(s',a;s)} \leq \sigma_{\mathcal{E}(s)}^2 \leq  \sum_{s',a,s'',a'}\sqrt{\sigma^2_{x(s',a;s)}\sigma^2_{x(s'',a';s)}},
\end{equation*}
where the upper bound is due to the Cauchy-Schwarz inequality, and the lower bound is since $\varepsilon(s,a)$ elements have a positive correlation (as reward trajectories overlap). Hence, it is evident that the improvement penalty can be extremely large when the term $\frac{|\beta-\pi|^2}{\beta}$ is unregulated and even a single mistake along the trajectory, caused by an unregulated element, might wreck the performance of the entire policy. However, by using the reroute constraint which tame each of these terms we can bound the variance of the improvement penalty.

While we analyzed the error for independent MC trajectories, a similar argument holds also for Temporal Difference (TD) learning \citep{Sutton2016ReinforcementL}. \citep{kearns2000bias} studied "bias-variance" terms in $k$-steps TD learning of the value function. Here we present their results for the Q-function error with TD updates. For any $0 < \delta < 1$, and a number $t$ of iteration through the data for the TD calculation, the maximal error term abides
\begin{equation}
\label{pac_td}
\varepsilon(s,a) \leq \max_{s,a}|\hat{Q}^{\beta}(s,a)-Q^{\beta}(s,a)|  
\leq \frac{1 - \gamma^{kt}}{1 - \gamma}\sqrt{\frac{3\log(k/\delta)}{N_s\beta(a|s)}} + \gamma^{kt}.    
\end{equation}
While the "bias", which is the second term in (\ref{pac_td}), depends on the number of iterations through the dataset, the "variance" which is the square root of the first term in (\ref{pac_td}) is proportional to $\frac{1}{\beta(a|s)N_s}$, therefore, bounding the ratio $\frac{|\beta-\pi|^2}{\beta}$ bounds the improvement penalty also for TD learning. 

\subsection{The Reroute Constraint}
In order to confine the ratio $\frac{|\beta-\pi|^2}{\beta}$, we suggest limiting the improvement step to a set of policies based on the following constraint.
\begin{definition}[Reroute Constraint]
Given a policy $\beta$, a policy $\pi$ is a $reroute(c_{\min},c_{\max})$ of $\beta$, if $\pi(a|s)=c(s,a)\beta(a|s)$ where $c(s,a)\in[c_{\min},c_{\max}]$. Further, note that reroute is a subset of the TV constraint with $\delta=\min(1-c_{\min}, \max(\frac{c_{\max}-1}{2},\frac{1-c_{\min}}{2}))$ \textit{(proof in the appendix)}.
\end{definition}
With reroute, each element in the sum of (\ref{improvement_penalty}) is proportional to $\sqrt{\beta(a|s)}|1-c(s,a)|$ where $c(s,a)\in[c_{\min},c_{\max}]$. Unlike reroute, other constraints such as the Total Variation (TV), forward and backward KL and PPO were not design to bound the improvement penalty. In the appendix, we analyze the improvement penalty under these constraints and show that it is generally unbounded.

\begin{algorithm}[]
\small{
\DontPrintSemicolon
\KwData{$s$, $\beta$, $A^{\beta}$, $(c_{\min}, c_{\max})$}
\KwResult{$\{\pi(a|s),\ a\in\mathcal{A}\}$}
\Begin{
$\mathcal{\tilde{A}} \longleftarrow \mathcal{A}$\;
$\Delta \longleftarrow 1 - c_{\min}$\;
$\pi(a|s) \longleftarrow c_{\min} \beta(a|s) \ \forall a \in \mathcal{A}$\;

\While{$\Delta > 0$}{
$a = \arg\max_{a\in\tilde{\mathcal{A}}} A^{\beta}(s,a)$\;
$\Delta_a = \min\{\Delta, (c_{\max} - c_{\min})\beta(a|s)\}$\;
$\mathcal{\tilde{A}} \longleftarrow \mathcal{\tilde{A}}/a$\;
$\Delta \longleftarrow \Delta - \Delta_a$\;
$\pi(a|s) \longleftarrow \pi(a|s) + \Delta_a$\;
}
}
}
\caption{Max-Reroute}
\label{alg:max_reroute}
\end{algorithm}

\subsection{Maximizing the Improvement Step under the Reroute Constraint}

We now turn to the problem of maximizing the objective function $J(\pi)$ under the reroute constraint and whether such maximization yields a positive improvement step. Maximizing the objective function without generating new trajectories of $\pi$ is a hard task since the distribution of states induced by the policy $\pi$ is unknown. Therefore, usually we maximize a surrogate off-policy objective function $J^{OP}(\pi) = \mathbb{E}_{s\sim\beta}[\sum_a\pi(a|s)A^{\beta}(s,a)]$. It is common to solve the constrained maximization with a NN policy representation and a policy gradient approach \citep{sutton2000policy, schulman2015trust}. Here we suggest an alternative: instead of optimizing a parametrized policy that maximizes $J^{OP}$, the actor (i.e. the agent that interact with the MDP environment) may ad hoc calculate a non-parametrized policy that maximizes the improvement step $\sum_a\pi(a|s)A^{\beta}(s,a)$ (i.e. the argument of the $J^{OP}$ objective) for each different state. This method maximizes also the $J^{OP}$ objective since the improvement step is independent between states. Note that with an ad hoc maximization, the executed policy is guaranteed to maximize the objective function under the constraint whereas with policy gradient methods one must hope that the optimized policy avoided NN caveats such as overfitting or local minima and converged to the optimal policy.  

For the reroute constraint, solving the non-parametrized problem amounts to solving the following simple linear program for each state 
\begin{equation}
\label{eq:max_reroute}
\begin{aligned}
\textrm{Maximize: } & (\bm{A}^{\beta})^T\bm{\pi} \\
\textrm{Subject to: } & c_{\min}\bm{\beta} \leq \bm{\pi} \leq c_{\max}\bm{\beta} \\
\textrm{And: } & \sum \pi_i = 1.
\end{aligned}
\end{equation}
Where $\bm{\pi}$, $\bm{\beta}$ and $\bm{A}^{\beta}$ are vector representations of $(\pi(a_i|s))_{i=1}^{|\mathcal{A}|}$, $(\beta(a_i|s))_{i=1}^{|\mathcal{A}|}$ and $(A^{\beta}(s,a))_{i=1}^{|\mathcal{A}|}$ respectively. We term the algorithm that solves this maximization problem as Max-Reroute (see Algorithm \ref{alg:max_reroute}). Similarly, one may derive other algorithms that maximize other constraints (see Max-TV and Max-PPO algorithms in the appendix). We will use Max-TV and Max-PPO as baselines for the performance of the reroute constraint. 

Notice that Max-Reroute, Max-TV, and Max-PPO satisfy the conditions of Corollary \ref{rank_base}, therefore they always provide a positive improvement step and hence, at least for a perfect approximation of $Q^{\beta}$ they are guaranteed to improve the performance. In addition, notice that they all use only the action ranking information in order to calculate the optimized policy. We postulate that this trait makes them more resilient to value estimation errors. This is in contrast to policy gradient methods which optimize the policy according to the magnitude of the advantage function.

\section{Two-armed bandit with Gaussian distributed rewards}
\label{bandit}
\begin{figure}[ht!]
  \centering
  \includegraphics[width=1\linewidth]{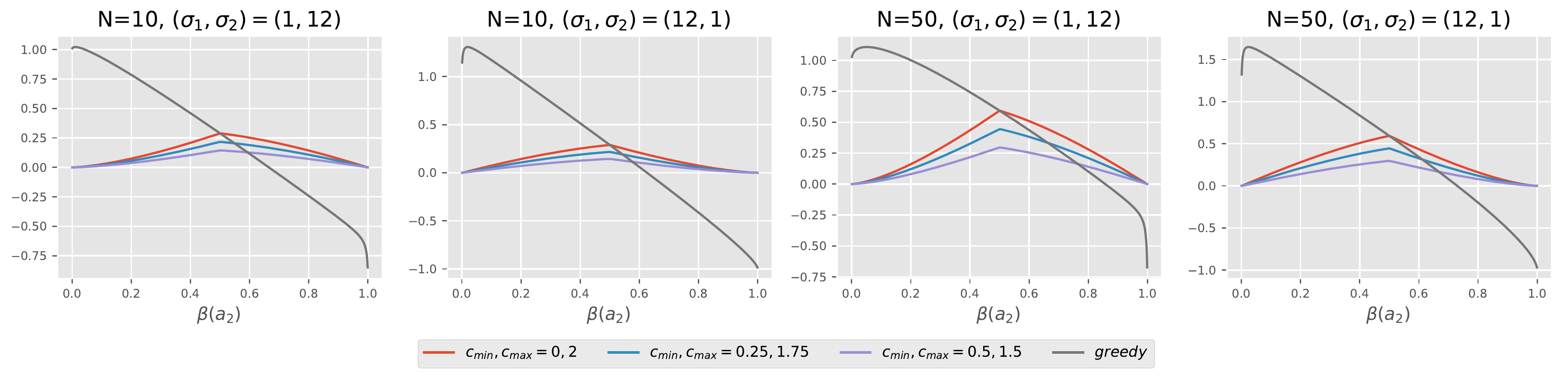}
  \caption{The negative regret with respect to a past behavior policy $\beta$. Negative number means that $\beta$ has a higher value than $\pi$.}
\label{fig:regret}
\end{figure}

To gain some insight into the nature of the RBI step, we begin by examining it in a simplified model of a two-armed bandit with Gaussian distributed rewards \citep{krause2011contextual}. To that end, define the reward of taking action $a_i$ as $r_i\sim\mathcal{N}(\mu_i, \sigma^2_i)$ and denote action $a_2$ as the optimal action s.t. $\mu_2 \geq \mu_1$. Let us start by considering the regret of a single improvement step from a batch of a past experience. In this case the regret is defined as
\begin{equation*}
    R^{\pi} = \mu_2 - V^{\pi} = \mu_2 - \sum_{i=1,2}\mathbb{E}[\pi(a_i)r_i] 
\end{equation*}
Specifically, we are interested in the difference between the behavior policy regret and the improved policy regret , i.e. $R^{\beta} - R^{\pi}$ which is equal to
\begin{equation*}
    R^{\beta} - R^{\pi} = P(I)V^{\pi}_{I} + (1 - P(I))V^{\pi}_{\bar{I}} -  V^{\beta}
\end{equation*}
where $I$ is an indicator of the clean event when $\hat{r}_2 > \hat{r}_1$ (where $\hat{r}_i$ is the empirical mean of $r_i$ over the batch data). Since $\hat{r}_1$ and $\hat{r}_2$ have Gaussian distributions, we can easily calculate $P(I)$. In Figure \ref{fig:regret} we plot the curves for $\mu_1=-1$ and $\mu_2=1$ for different number of batch sizes and different variances. As a baseline we plot the difference in regret for a greedy step. First, we see that for poor behavior policies s.t. $\beta(a_2) \leq \beta(a_1)$, it is better to make a greedy step. This is because the suboptimal action has sufficiently accurate estimation. However for good behavior policies, s.t. $\beta(a_2) > \beta(a_1)$ (which is generally the case when learning from demonstrations and even in RL after the random initialization phase), the greedy regret grows significantly s.t. it is better to stick with the behavior policy. This is obvious since while the behavior policy reaches to the optimum, a wrong value evaluation of the inferior action triggers a bad event $\bar{I}$ which leads to a significant degradation. On the other hand, a RBI step demonstrates safety (i.e. lower regret than $\beta$) and no less important, its regret is lower than the greedy regret for good behavior policies. This pattern repeats for all reroute parameters s.t. $c_{\min} \leq 0.5$, $c_{\max} \leq 2$. \footnote{For large enough $c_{\max}$ there is a potential for an unsafe policy, specifically for $c_{\min}=0$ and $c_{\max}\to \infty$ the reroute constraint converges to a greedy step.}

\begin{figure}[ht!]
  \centering
  \includegraphics[width=1\linewidth]{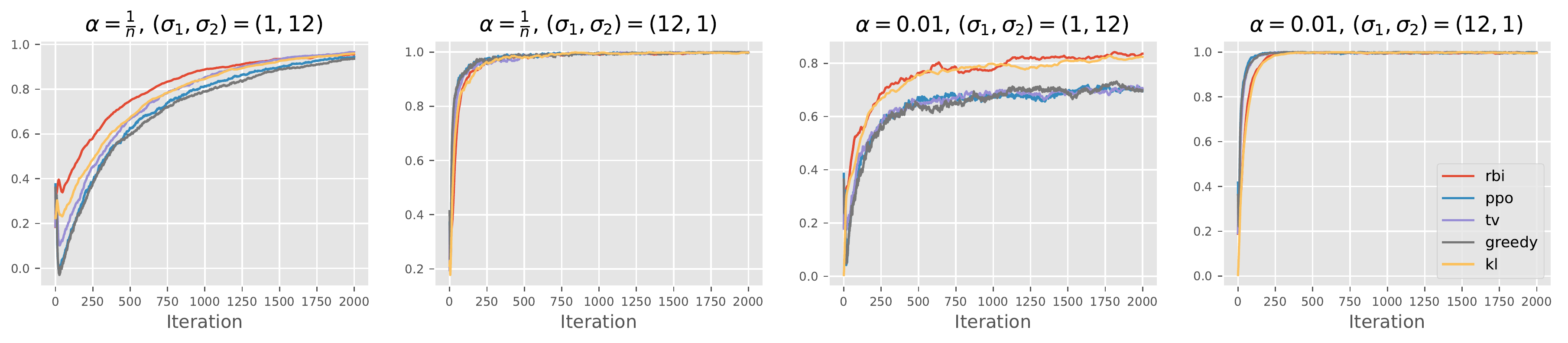}
  \caption{Different constrained policies performances in a two-armed bandit with Gaussian distributed reward setting}
\label{fig:bandit}
\end{figure}

To find how RBI performs in an iterative RL process, we continue with the bandit example but now we consider the learning curve of off-policy learning where the behavior policy is mixed with a fix exploration parameter, i.e. $\beta(a)=\pi(a)(1 - \varepsilon) + \frac{\varepsilon}{n_a}$ (where $n_a$ is the number of actions and $\varepsilon=0.1$). The Q-function is learned with $Q^{\pi}(a) = (1 - \alpha) Q^{\pi}(a) + \alpha r$, where $\alpha$ is a learning rate, possibly decaying over time. We evaluate several constrained policies: (1) RBI with $(c_{\min}, c_{\max})=(0.5, 1.5)$, (2) PPO with $\varepsilon=0.5$, (3) TV with $\delta=0.25$, (4) greedy step and; (5) forward KL with $\lambda=1$. RBI, TV and PPO were all maximized with our maximization algorithms (without gradient ascent). To avoid absolute zero probability actions, we clipped the policy such that $\pi(a_i) \geq 10^{-3}$. In addition we added 10 random sample at the start of the learning process. The learning curves are plotted in Figure \ref{fig:bandit}. 

The learning curves exhibit two different patterns. For the scenario of $\sigma_1 > \sigma_2$, a fast convergence of all policies was obtained. Essentially, when the better action has low variance it is easy to discard the worse action by choosing it and rapidly improving its value estimation and then switching to the better action. On the other hand, for the case of $\sigma_1 < \sigma_2$ it is much harder for the policy to improve the estimation of the better action after committing to the worse action. We see that RBI defers early commitment and while it slightly reduces the rate of convergence in the first (and easy) scenario, it significantly increases the data efficiency in the harder scenario. We postulate that the second scenario is also more important to real-world problems where usually one must take a risk to get a big reward (where avoiding taking risks, significantly reduces the reward variance but also the expected reward).

In the second scenario, RBI has the best and KL has the second-best learning curves in terms of initial performance. However, there is another distinction between the ideal learning rate (LR) of $\alpha=\frac{1}{n}$ and a constant rate of $\alpha=0.01$.  In the ideal LR case, the advantage of RBI and KL reduces over time. This is obvious since a LR of $\alpha=\frac{1}{n}$ takes into account the entire history and as such, for large history, after a large number of iterations, there is no need for a policy which learns well from a finite dataset. On the other hand, there is a stable advantage of RBI and KL for a fix LR as fix LR does not correctly weight the entire past experience. Notice that in a larger than 1-step MDP, it is unusual to use a LR of $\frac{1}{n}$ since the policy changes as the learning progress, therefore, usually the LR is fixed or decays over time (but not over state visitations). Hence, the advantage of RBI over greedy should be realized through the entire training process. 

\section{Learning to play Atari by observing human players}

In section \ref{policy_improvement}, the reroute constraint was derived for learning from a batch of past experience of a single behavior policy $\beta$ and a tabular value and policy parametrization. In this section, we empirically extend our results to a NN parametric form (both for value and policy) and an experience dataset of multiple policies $\{\beta_i\}$. To that end, we use a crowdsourced dataset of 4 Atari games (SpaceInvaders, MsPacman, Qbert and Montezuma's Revenge) \citep{kurin2017atari}, where each game has roughly 1000 recorded episodes. Such a dataset of observations is of a particular interest since it is often easier to collect observations than expert demonstrations. Therefore, it may be practically easier to initialize a RL agent with learning from observations than generate trajectories of a single expert demonstrator.

For our purpose, we simply aggregated all trajectories into a single dataset and estimated the average policy $\beta$ as the average behavior of the dataset. Practically, when using a policy network, this sums to simply minimize the Cross-Entropy loss between the $\beta_{\theta}(\cdot|s)$ and the empirical evidence $(s, a)$ as in a plain classification problem. To estimate the $Q$-value of $\beta$ we employed a Q-network in the form of the Dueling DQN architecture \citep{mnih2015human,wang2015dueling} and minimized the MSE loss between Monte-Carlo reward trajectories $\sum_{k \geq 0}\gamma^kr_k$ and the $Q^{\beta}(s,a)$ estimation. Note that this does not necessarily converge to the true $Q$-value of $\beta$ (since the trajectories were generated by the behavior policies $\{\beta_i\}$) but it empirically provides a close estimation with a small error. 

Given our estimations of $\beta$ and $Q^{\beta}$, we applied several improvement steps and evaluated their performance. For the RBI, TV and greedy steps, we calculated the improved policy $\pi$ in the non-parametric space (i.e. during evaluation) with Max-Reroute and Max-TV, for the PPO step we applied additional gradient ascent optimization step and calculated the PPO policy. As an additional baseline, we implemented the DQfD algorithm with hyperparameters as in \cite{hester2018deep}. 

\begin{figure}[ht!]
  \centering
  \includegraphics[width=1\linewidth]{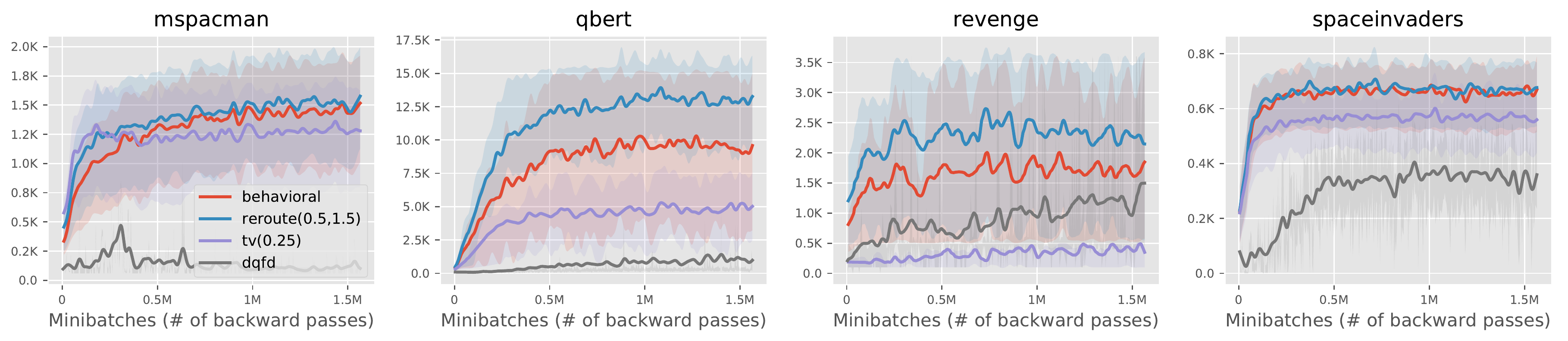}
  \caption{Learning to play Atari from a dataset of human players: Learning curves}
\label{fig:lfo_atari_resluts}
\end{figure}

\begin{table}[ht]
\label{lfo_table}
\centering
\caption{Final scores of different policies}
\begin{tabular*}{8cm}[t]{lrrrr}
\toprule
Method & Pacman & Qbert & Revenge & SI \\
\midrule
 Humans               &   3024    & 3401    &   1436  &   634\\
 Behavioral   &   1519    & 8562    &  1761  &   678\\
 RBI$(0.5,1.5)$  &   \textbf{1550}    & 12123   &  \textbf{2379}  &   \textbf{709}\\
 TV$(0.25)$           &   1352    & 5089    &  390   &   573\\
 PPO$(0.5)$           &   1528    & \textbf{14089}   &  388   &   547\\
 DQfD                 &   83      & 1404    &  1315  &   402\\
\bottomrule
\end{tabular*}
\end{table}

First, we found out that behavioral cloning, i.e. simply playing with the calculated average behavior $\beta$, yielded generally good results with the exception of MsPacman which is known to be a harder game (see comparison of human and DQN scores in \cite{mnih2015human}). For Qbert, the behavioral score was much better than the average score and we assume that this is since good episodes tend to be longer in Qbert, hence their weight in the average behavior calculation is greater.

We evaluated Max-TV with $\delta=0.25$ since it encapsulates the $reroute(0.5, 1.5)$ region. However, unlike $reroute(0.5, 1.5)$ a TV constrained step reduced the performance below the behavioral cloning performance in all games. At a first glance, this may be a surprising evidence but it is expected after analyzing Eq. (\ref{improvement_penalty_eq}). Only a single state with a bad estimation along the trajectory is required to significantly reduce the overall score. On the other hand, $reroute(0.5,1.5)$ always increased the behavioral score and generally provided the best performance.

While PPO with $\varepsilon=0.5$ scored generally better than TV, we noticed that in two games it reduced the behavioral score. Our extended results (see appendix) showed similarity between $PPO(0.5)$ and $reroute(0,2)$. This indicates that, PPO tends to settle negative advantage action to zero probability in order to avoid the negative PPO penalty. Empirically, this also indicates that it is important to set $c_{\min} > 0$ to avoid reducing the probability of good actions to zero due to value estimation errors. Finally, we observed the significantly lower score of the DQfD algorithm. DQfD uses the demonstrations data to learn a policy with an almost completely off-policy algorithm (it adds a regularization term that should tie the learned policy to the expert behavior, but its effectiveness depends on the $Q$-value magnitude). This strengthen our assertion that when learning from a fixed size past experience, the calculated policy must be constrained to the behavior policy presented in the data.

\section{Distributed RL with RBI}

We now turn to implementation of RBI learning in an iterative RL setting in the Atari environment without utilizing any prior human demonstrations. We adopted a distributed learning setting with multiple actors and a single learner process, similar to the setting of Ape-X \cite{horgan2018distributed}. However, contrary to the Ape-X algorithm, which learns only a single $Q$-value network (and infers the policy with a greedy action selection), we learned side-by-side two different networks: (1) a $Q$-value network, termed $Q$-net and denoted $Q^{\pi}_{\phi}$ and; (2) a policy network termed $\pi$-net and denoted $\pi_{\theta}$.

In this experiment we attempt to verify: (1) whether RBI is a good approach in DRL also for better final performance (a tabular example of RL with RBI was discussed in section \ref{bandit}) and (2) whether our approach of solving for the optimal policy in the non-parametrized space as part of the actor's routine, can be generalized to iterative RL. Note that recently, \cite{vuong2018supervised} developed a framework of constrained RL where they generate samples with a policy $\pi_{\theta_k}$ (where $\theta_k$ is the set of network parameters in the $k$-th learning step), and in the learner process, they calculate an improved non-parameterized policy $\pi$ under a given constrained and learn the next policy $\pi_{\theta_{k+1}}$ by minimizing the KL distance $D_{KL}(\pi_{\theta_{k+1}}, \pi)$. 

We adopted a slightly different approach. Instead of calculating the non-parameterized policy in the learner process, each actor loads a stored policy $\pi_{\theta_k}$ and calculates the non-parameterized optimal policy $\pi$ as part of the actor routine while interacting with the environment and generating its trajectories. The optimized policy for each new state is then stored to a replay buffer and executed by the actor (after mixing with a small random exploration). The learner process learns to imitate the optimized policy by minimizing the loss $D_{KL}(\pi,\pi_{\theta_{k+1}})$. This approach has several additional benefits:
\begin{enumerate}
    \item The learner is not required to calculate the optimized policy, therefore, the rate of processing minibatches of past experience increases. This rate is generally the bottleneck of the wall clock time of the process.
    \item The $Q$-net expresses the estimation of the value of the $\pi$-net policy. In contrary to \cite{vuong2018supervised} where the $Q$-net describes the value of the past policy but the $\pi$-net is the future policy. This has benefit of a more accurate value estimation which is necessary to calculate target updates and prioritization weights. 
    \item Our executed policy $\pi$ is the optimal solution of the constraint and there is no need to wait for the network to converge to the non-parameterized solution.
    This also helps when the optimal policy has high non-linearity with respect to the input state.
\end{enumerate}

We found out that good RBI parameters that work well and balance between safety and fast convergence are $(c_{\min},c_{\max})=(0.1, 2)$. We also found out that it is better to mix the RBI solution with a small amount of a greedy policy for a faster recovery of actions with near-zero probability. Therefore, the actor policy was
\begin{equation*}
    \pi = (1 - c_{greedy}) \pi^{rbi} + c_{greedy} \pi^{greedy},
\end{equation*}
with $c_{greedy}=0.1$. Finally, as for Q-learning with prioritization \cite{schaul2015prioritized}, we noticed that a significant improvement is achieved when adding a prioritization to the policy loss function.\footnote{Note that, for simplicity, we did not use a priority replay, rather we just weighted the loss by the priority weights.} We hypothesized that $\pi$-net should focus on states in which a small error leads to a large value difference. To emphasize such states we applied weights based on the advantage variance over the policy $\pi_{\theta}$ mixed with a random noise, i.e. $\pi^{mix} = (1-c_{mix})\pi_{\theta} + c_{mix}\pi^{rand}$. The priority is therefore
\begin{equation}
\label{adv_var}
w^{\pi} \propto  \left(\frac{\sum_{a}\pi^{mix}(a|s)(A^{\pi}_{\phi}(a,s))^2 + 0.1}{|V^{\pi}_{\phi}| + 0.1}\right)^{\alpha}.
\end{equation} 
Where $\alpha$ is a damping factor. All other parameters in our environments, such as exploration rates and the optimizer parameters were identical between the Ape-X baseline and the RBI learner. The final algorithms for the learner and actors are presented in Algorithms \ref{alg:learner} and \ref{alg:actor}.

\begin{algorithm}[ht!]
\small{
\DontPrintSemicolon
\KwData{$\mathcal{D}=\{(s_i,a_i,\pi_i,r_i,\delta_i),..\}$}
\KwResult{$\theta$, $\phi$}
\While {True}{
\BlankLine
$B \longleftarrow \textrm{PriorityBatchSample}(\mathcal{D}, N)$\;
$\pi_{\theta_k}\longleftarrow \pi\textrm{-net}(s)$\;
$Q^{\pi}_{\phi_k}\longleftarrow Q\textrm{-net}(s, a, \cdot)$\;
Calculate a target value with a target network:\;
$R_i=\sum_{t=1}^n\gamma^t r_{i+t} + \gamma^{n}V^{\pi}_{\bar{\phi}}(s_{i+n})$\;
$\mathcal{L}^{\pi}(\pi_{\theta_k})= \frac{1}{N}\sum_{i\in B} w_i D_{KL}\left( \pi_i, \pi_{\theta_k, i} \right)$ 
\BlankLine
$\mathcal{L}^{q}(Q^{\pi}_{\phi_k})=\frac{1}{N}\sum_{i\in B} w_i (R_i - Q^{\pi}_{\phi_k, i})^2$\;
\BlankLine
$\theta_{k+1} = \theta_{k} - \alpha \nabla_{\theta} \mathcal{L}^{\pi}(\pi_{\theta_k})$\;
\BlankLine
$\phi_{k+1} = \phi_{k} - \alpha \nabla_{\phi} \mathcal{L}^{q}(Q^{\pi}_{\phi_k})$
}
}
\caption{RBI learner}
\label{alg:learner}
\end{algorithm}

\begin{algorithm}[ht!]
\small{
\DontPrintSemicolon
$s\longleftarrow\textrm{ResetMDP()}$\;
\While{t = 0}{
$\hat{\pi}\longleftarrow \pi\textrm{-net}(s)$\;
$Q^{\hat{\pi}}, V^{\hat{\pi}} \longleftarrow Q\textrm{-net}(s, \cdot, \hat{\pi})$\;
$\pi\longleftarrow \textrm{MaxReroute}(\hat{\pi},s, A^{\hat{\pi}})$\;
\BlankLine
$\beta \longleftarrow \textrm{AddExploration}(\pi)$\;
$a \longleftarrow \textrm{Sample}(\beta )$\;
$s,r,t \longleftarrow \textrm{StepMDP}(a)$\;
}
$\delta \longleftarrow \textrm{CalcPriorities}(\{r\}, \{V^{\hat{\pi}}\}, \gamma)$\;
\BlankLine
$\textrm{AddToDataset}(\{(s_i,a_i,\pi_i,r_i,\delta_i),...\})$\;
}
\caption{RBI actor}
\label{alg:actor}
\end{algorithm}

\begin{figure}[ht!]
  \centering
  \includegraphics[width=1\linewidth]{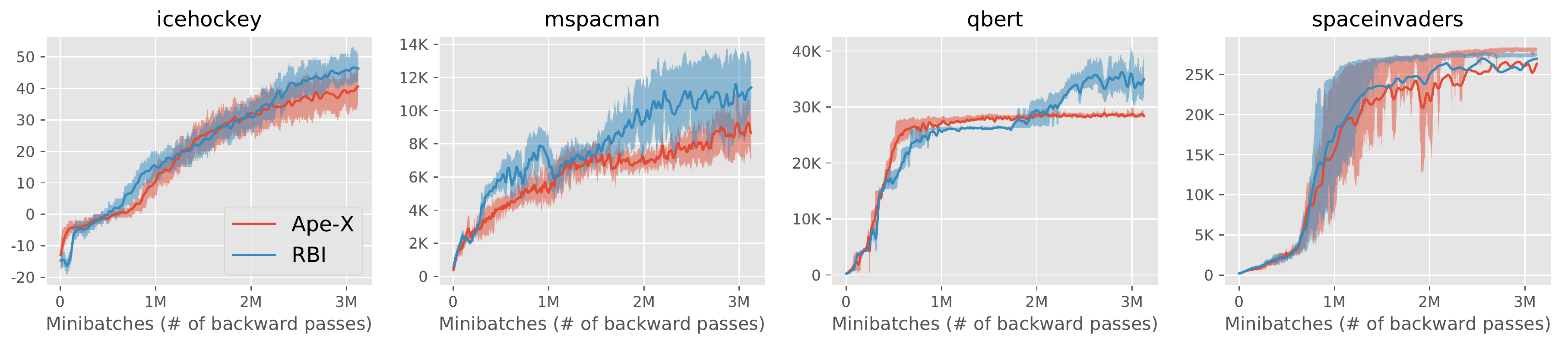}
  \caption{Performance curves of 4 Atari games. The second and third quartiles are shadowed.}
\label{fig:rl_atari_resluts}
\end{figure}

We evaluated RBI and the Ape-X baseline over a set of 12 Atari games with a varying range of difficulty. As an external baseline we compared our results to the Rainbow score \cite{hessel2017rainbow}. For a fair comparison, we used a batch size of 128 and capped the learning process to $N = 3.125M$ backward passes. This is identical to the number of different states that were processed by the network in Rainbow (12.5M backward passes with a batch of 32). Altogether, RBI provided better policies in 7 out of 11 environments with respect to our Ape-X implementation and in 9 out of 11 environments with respect to the Rainbow score (all tied in Freeway). See table \ref{table:rbi_score} and extended results in the appendix. In some games (e.g. Qbert) RBI improved its policy while Ape-X converged to a steady policy. We postulate that with NN parametrization, taking greedy steps may lead to a faster convergence of the nets since the diversity of actions is much lower. Generally, high diversity, i.e. a higher entropy policy is known to be beneficial for exploration \cite{haarnoja2018soft} and indeed we noticed that RBI learns faster hard exploration tasks such as oxygen replenish in Seaquest. Overall, we found that only in a single environment (Berzerk) sticking with the greedy policy paid off considerably, and even in this case for the majority of the learning process RBI was superior.

\section{Conclusions}

We introduced a constrained policy improvement method, termed RBI, designed to learn a safe policy from a batch of past experience. In addition to safety, we found that in a two-armed bandit with Gaussian distributed reward, RBI is more data efficient than greedy and other constrained policies. We also introduced our approach to train constrained policies such as RBI. An actor solves a non-parametrized reroute constrained optimization problem (Eq \ref{eq:max_reroute}) and a learner imitates the actor's policy with a parametrized network and a KL divergence loss. Our initial results demonstrate a gain in data efficiency and final performance when using RBI instead of Q-learning greedy approaches.

\newpage
\bibliography{mybib}
\bibliographystyle{iclr2019_conference}
\newpage

\onecolumn
\appendix

\appendix

\section{Rerouted Behavior Improvement Proofs}
\label{policy_improvement_proofs}

\subsection{Soft Policy Improvement}
The soft policy improvement rule states that every policy $\pi$ that satisfies
\begin{equation*}
\sum_{a}\pi(a|s)A^{\beta}(s,a) \geq 0 \ \ \forall s \in \tilde{\mathcal{S}}
\end{equation*}
improves the policy $\beta$ s.t. $V^{\pi} \geq V^{\beta} \ \forall s$. To avoid stagnation we demand that the inequality be strict for at least a single state. In operator notation, the last equation can be written as
\begin{equation*}
\Pi^{\pi}a^{\beta} \geq 0
\end{equation*}
\begin{proof}
Plugging in $a^{\beta}=q^{\beta}-v^{\beta}$ and using $v^{\beta}=\Pi^{\pi}v^{\beta}$, we get that
\begin{equation}
\label{improvement_operator}
v^{\beta} \leq \Pi^{\pi}q^{\beta} = \Pi^{\pi}(r + \gamma \tilde{\mathcal{P}} v^{\beta}) = r^{\pi} + \gamma P_v^{\pi} v^{\beta}.
\end{equation}
Then, by applying (\ref{improvement_operator}) recursively we get
\begin{multline*}
v^{\beta} \leq r^{\pi} + \gamma P_v^{\pi} v^{\beta} = r^{\pi} + \gamma P_v^{\pi} (r^{\pi} + \gamma P_v^{\pi} v^{\beta}) \leq r^{\pi} + \gamma P_v^{\pi}r^{\pi} + \gamma^2 (P_v^{\pi})^2v^{\beta} \leq r^{\pi} + \\
\gamma P_v^{\pi}r^{\pi} + \gamma^2 (P_v^{\pi})^2r^{\pi} + \gamma^3 (P_v^{\pi})^3v^{\beta}
\leq ... \leq \sum_{k \geq 0} \gamma^{k}(P_v^{\pi})^k r^{\pi} = (\mathcal{I} - \gamma P_v^{\pi})^{-1} r^{\pi} = v^{\pi}.
\end{multline*}
\end{proof}
\citep{schulman2015trust} have shown that
\begin{equation*}
J(\pi) - J(\beta) = \mathbb{E}_{T\sim\pi}\left[\sum_{k \geq 0}\gamma^{k}A^{\beta}(s_k,a_k)\right].
\end{equation*}
This equation can also be written as
\begin{multline*}
J(\pi) - J(\beta) = \sum_{k \geq 0}\gamma^{k}\mathbb{E}_{s_k\sim\pi|s_0}\left[\sum_{a\in\mathcal{A}}\pi(a|s_k)A^{\beta}(s_k,a_k)\right] = \\
=\sum_{k \geq 0}\gamma^{k}\sum_{s\in\mathcal{S}}P(s_0 \xrightarrow[]{k} s| \pi)\sum_{a\in\mathcal{A}}\pi(a|s)A^{\beta}(s,a) = \\
\sum_{s\in\mathcal{S}}\left(\sum_{k \geq 0}\gamma^{k}P(s_0 \xrightarrow[]{k} s| \pi)\right)\sum_{a\in\mathcal{A}}\pi(a|s)A^{\beta}(s_k,a).
\end{multline*}
Recall the definition of the discounted distribution of states
\begin{equation*}
\rho^{\pi}(s) = \sum_{k \geq 0} \gamma^k P(s_0 \xrightarrow[]{k} s| \pi),
\end{equation*}
we conclude that
\begin{equation*}
J(\pi) - J(\beta) = \sum_{s\in\mathcal{S}}\rho^{\pi}(s)\sum_{a\in\mathcal{A}}\pi(a|s)A^{\beta}(s,a)
\end{equation*}
\subsection{Rank-Based Policy Improvement}

\begin{proof}

Denote by $i$ the index of the advantage ordered list $\{A_i\}$. Since $\sum_{a_i}\beta_iA^{\beta}_i = 0 $, we can write it as an equation of the positive and negative advantage components
\begin{equation*}
\sum_{a=i_p}^{N}\beta_iA^{\beta}_i = \sum_{a=0}^{i_n}\beta_i(-A^{\beta}_i),
\end{equation*}
where $i_p$ is the minimal index of positive advantages and $i_n$ is the maximal index of negative advantages. Since all probability ratios are non-negative, it is sufficient to show that
\begin{equation*}
\sum_{a=i_p}^{N}c_i\beta_iA^{\beta}_i \geq \sum_{a=0}^{i_n}c_i\beta_i(-A^{\beta}_i).
\end{equation*}
But clearly
\begin{equation*}
\sum_{a=i_p}^{N}c_i\beta_iA^{\beta}_i \geq c_{i_p}\sum_{a=i_p}^{N}\beta_iA^{\beta}_i \geq c_{i_n}\sum_{a=0}^{i_n}\beta_i(-A^{\beta}_i) \geq \sum_{a=0}^{i_n}c_i\beta_i(-A^{\beta}_i).
\end{equation*}

\end{proof}

\subsection{Policy Improvement Penalty}

For the proof, we use vector and operator notation, where $v^{\pi}$ and $q^{\pi}$ represent the value and $Q$-value functions as vectors in $\mathcal{V}\equiv\mathbb{R}^{|\mathcal{S}|}$ and $\mathcal{Q}\equiv\mathbb{R}^{|\mathcal{S}\times\mathcal{A}|}$. We denote the partial order $x \geq y \iff x(s,a) \geq y(s,a) \ \ \forall s,a\in \mathcal{S}\times\mathcal{A}$, and the norm $\|x\|=\max_{s,a}|x|$. Let us define two mappings between $\mathcal{V}$ and $\mathcal{Q}$:
\begin{enumerate}
\item 
$\mathcal{Q}\to\mathcal{V}$ mapping $\Pi^{\pi}$:  $\left(\Pi^{\pi}q^{\pi}\right)(s)=\sum_{a\in\mathcal{A}}\pi(a|s)q^{\pi}(s,a)=v^{\pi}(s)$. \\
This mapping is used to backup state-action-values to state value.
\item 
$\mathcal{V}\to\mathcal{Q}$ mapping $\tilde{\mathcal{P}}$:  $\left(\tilde{\mathcal{P}}v^{\pi}\right)(s,a)=\sum_{s'\in\mathcal{S}}P(s'|s,a)v^{\pi}(s')$. \\
This mapping is used to backup state-action-values based on future values. Note that this mapping is not dependent on a specific policy.
\end{enumerate}
Let us further define two probability operators: \begin{enumerate}
\item 
$\mathcal{V}\to\mathcal{V}$ state to state transition probability $P_v^{\pi}$:  \\ 
$\left(P_v^{\pi}v^{\pi}\right)(s)=\sum_{a\in\mathcal{A}}\pi(a|s)\sum_{s'\in\mathcal{S}}P(s'|s,a)v^{\pi}(s')$. 
\item 
$\mathcal{Q}\to\mathcal{Q}$ state-action to state-action transition probability $P_q^{\pi}$: \\  
$\left(P_q^{\pi}q^{\pi}\right)(s,a)=\sum_{s'\in\mathcal{S}}P(s'|s,a)\sum_{a'\in\mathcal{A}}\pi(a'|s')q^{\pi}(s', a')$.
\end{enumerate}
Note that the probability operators are bounded linear transformations with spectral radius $\sigma(P_v^{\pi})\leq 1$ and $\sigma(P_q^{\pi})\leq 1$ \citep{puterman2014markov}.
Prominent operators in the MDP theory are the recursive value operator $T_v^{\pi}v=\Pi^{\pi}r+\gamma P_v^{\pi}v$ and recursive $Q$-value operator $T_q^{\pi}q=r+\gamma P_q^{\pi}q$. Both have a unique solution for the singular value equations $v=T_v^{\pi}v$ and $q=T_q^{\pi}q$. These solutions are $v^{\pi}$ and $q^{\pi}$ respectively \citep{Sutton2016ReinforcementL,puterman2014markov}. In addition, in the proofs we will use the following easy to prove properties:
\begin{enumerate}
\item
$P_q^{\pi}=\tilde{\mathcal{P}}\Pi^{\pi}$
\item
$P_v^{\pi}=\Pi^{\pi}\tilde{\mathcal{P}}$
\item
$x \geq y, \ \ x,y\in\mathcal{V}, \ \Rightarrow \tilde{\mathcal{P}}x \geq \tilde{\mathcal{P}}y$
\item
The probability of transforming from $s$ to $s'$ in $k$ steps is $P(s \xrightarrow[]{k} s'|\pi) = (P_v^{\pi})^k(s,s')$.
\item
$[(P_v^{\pi})^k x](s) = \mathbb{E}_{s_k\sim\pi|s}\left[x(s_k)\right]$
\end{enumerate}

Let $\hat{Q}^{\beta}$ be an approximation of $Q^{\beta}$ with an error $\varepsilon(s,a) = (Q^{\beta} - \hat{Q}^{\beta})(s,a)$ and let $\pi$ be a policy that satisfies the soft policy improvement theorem with respect to $\hat{Q}^{\beta}$. Then the following holds:
\begin{equation}
\label{value_penalty}
V^{\pi}(s) - V^{\beta}(s) \geq  - \sum_{s'\in\mathcal{S}}\left(\sum_{k \geq 0} \gamma^k P(s\xrightarrow[]{k}s'|\pi)\right)\sum_{a\in\mathcal{A}}\varepsilon(s',a)\left(\beta(a|s')-\pi(a|s')\right).
\end{equation}
The proof resembles the Generalized Policy improvement theorem \citep{barreto2017successor}.

\begin{proof}
We will use the equivalent operator notation. Define the vector $\varepsilon = q^{\beta} - \hat{q}^{\beta}$
\begin{equation*}
\begin{split}
T_q^{\pi}\hat{q}^{\beta}  = r + \gamma\tilde{\mathcal{P}}\Pi^{\pi}\hat{q}^{\beta} & \geq r + \gamma\tilde{\mathcal{P}}\Pi^{\beta}\hat{q}^{\beta}\\
 & = r + \gamma\tilde{\mathcal{P}}\Pi^{\beta}q^{\beta} - \gamma\tilde{\mathcal{P}}\Pi^{\beta}\varepsilon \\
 & = T_q^{\beta}q^{\beta} - \gamma P_q^{\beta} \varepsilon \\ 
 & = q^{\beta} - \gamma P_q^{\beta} \varepsilon \\ 
 & = \hat{q}^{\beta} + (1 - \gamma P_q^{\beta}) \varepsilon \\ .
\end{split}
\end{equation*}
Note that the inequality is valid, since $\Pi^{\pi}\hat{q}^{\beta} \geq \Pi^{\beta}\hat{q}^{\beta}$ (by the theorem assumptions), and if $v \geq u$ then $\tilde{\mathcal{P}}v \geq \tilde{\mathcal{P}}u$.
Set $y = (1 - \gamma P_q^{\beta}) \varepsilon$ and notice that $T_q^{\pi}(\hat{q}^{\beta}+y)=T_q^{\pi}\hat{q}^{\beta} + \gamma P_q^{\pi} y$. By induction, we show that
\begin{equation*}
(T_q^{\pi})^n \hat{q}^{\beta} \geq \hat{q}^{\beta} + \sum_{k=0}^n \gamma^k (P_q^{\pi})^k y.
\end{equation*}
We showed it for $n=1$, assume it holds for $n$, then for $n+1$ we obtain
\begin{multline*}
(T_q^{\pi})^{n+1} \hat{q}^{\beta} = T_q^{\pi} (T_q^{\pi})^n \hat{q}^{\beta} \geq T_q^{\pi} \left( \hat{q}^{\beta} + \sum_{k=0}^n \gamma^k (P_q^{\pi})^k y \right)= \\
= T_q^{\pi}\hat{q}^{\beta} + \sum_{k=0}^n \gamma^{k+1} (P_q^{\pi})^{k+1} y \geq \hat{q}^{\beta} + y +  \sum_{k=0}^n \gamma^{k+1} (P_q^{\pi})^{k+1} = 
\hat{q}^{\beta} + \sum_{k=0}^{n+1} \gamma^k (P_q^{\pi})^k y.
\end{multline*}

Using the contraction properties of $T_q^{\pi}$, s.t. $\lim_{k\to\infty}(T_q^{\pi})^k x = q^{\pi}$, $\forall x \in \mathcal{Q}$ and plugging back $(1 - \gamma P_q^{\beta}) \varepsilon = y$, we obtain

\begin{equation*}
\begin{split}
q^{\pi}  = \lim_{n\to\infty}(T_q^{\pi})^n(\hat{q}^{\beta}) & \geq \hat{q}^{\beta} + \sum_{k\geq 0} \gamma^k(P_q^{\pi})^k (1 - \gamma P_q^{\beta}) \varepsilon \\
 & = \hat{q}^{\beta} + \varepsilon + \sum_{k\geq 0} \gamma^{k+1}(P_q^{\pi})^k (P_q^{\pi} - P_q^{\beta}) \varepsilon.
\end{split}
\end{equation*}
Applying $\Pi^{\pi}$ we transform back into $\mathcal{V}$ space
\begin{equation*}
\begin{split}
v^{\pi} & \geq \Pi^{\pi}\hat{q}^{\beta} + \Pi^{\pi}\varepsilon + \sum_{k\geq 0} \gamma^{k+1}\Pi^{\pi}(P_q^{\pi})^k (P_q^{\pi} - P_q^{\beta}) \varepsilon  \\
& \geq \Pi^{\beta}\hat{q}^{\beta} + \Pi^{\pi}\varepsilon + \sum_{k\geq 0} \gamma^{k+1}\Pi^{\pi}(P_q^{\pi})^k (P_q^{\pi} - P_q^{\beta}) \varepsilon \\ 
& = v^{\beta} + (\Pi^{\pi} - \Pi^{\beta})\varepsilon + \sum_{k\geq 0} \gamma^{k+1}\Pi^{\pi}(P_q^{\pi})^k (P_q^{\pi} - P_q^{\beta}) \varepsilon  \\.
\end{split}
\end{equation*}
Notice that $\Pi^{\pi}(P_q^{\pi})^kP_q^{\pi} = \Pi^{\pi}(\tilde{\mathcal{P}}\Pi^{\pi})^k \tilde{\mathcal{P}}\Pi^{\pi} = (\Pi^{\pi}\tilde{\mathcal{P}})^{k+1}\Pi^{\pi}=(P_v^{\pi})^{k+1}\Pi^{\pi}$, and in the same manner, $\Pi^{\pi}(P_q^{\pi})^kP_q^{\beta} = (P_v^{\pi})^{k+1}\Pi^{\beta}$. Therefore, we can write
\begin{equation*}
v^{\pi} \geq v^{\beta} - \sum_{k \geq 0} \gamma^k (P_v^{\pi})^{k}(\Pi^{\beta} - \Pi^{\pi})\varepsilon,
\end{equation*}
which may also be written as (\ref{value_penalty}).
\end{proof}

\subsection{Reroute is a subset of TV}

The set of reroute policies with $[c_{\min},c_{\max}]$ is a subset of the set of $\delta$-TV policies, where $\delta=\min(1-c_{\min}, \max(\frac{c_{\max}-1}{2},\frac{1-c_{\min}}{2}))$.
\begin{proof}
For a reroute policy, the TV distance is $\frac{1}{2}\sum_{a}\beta(a_i|s)|1-c_i|\leq \sum_{a}\beta(a_i|s)\delta =  \delta$. For the second upper bound $1-c_{\min}$, notice that a rerouted policy $\{\pi_i\}$ may be written as $\pi_i=c_{\min}\beta_i+\delta_i$, where $\delta_i\geq 0$. Therefore, the TV distance is
\begin{equation*}
TV = \frac{1}{2}\sum_{i}|\pi_i - \beta_i| = \frac{1}{2}\sum_{i}|c_{\min}\beta_i+\delta_i - \beta_i| \leq \frac{1}{2}\sum_{i}\left((1 - c_{\min})\beta_i+\delta_i\right),
\end{equation*}
but $1 = \sum_i \pi_i = \sum_i(c_{\min}\beta_i+\delta_i) = c_{\min} + \sum_i \delta_i$. Thus, $\sum_i \delta_i = 1 - c_{\min}$ and we may write
\begin{equation*}
TV  \leq \frac{1}{2} \left((1 - c_{\min}) + (1 - c_{\min})\right) = 1 - c_{\min}.
\end{equation*}

To show that the TV region is not in reroute, take any policy with a zero probability action and another action with a probability of $\beta(a_i|s) \leq \delta$. The policy which switches probabilities between these two actions is  in $TV(\delta)$ but it is not in reroute for any finite $c_{\max}$.
\end{proof}

\subsection{Analyzing other policy improvement algorithms}

Let us now consider other popular constraints. Clearly, with TV $\delta \geq \frac{1}{2}\sum_{a_i}|\beta_i-\pi_i|$ the ratio $\frac{|\beta-\pi|^2}{\beta}$ is unregulated since actions with a zero probability may increase their probability up to $\delta$. For the PPO constraint, we prove two important characteristics, first, it does not regulate the ratio $\frac{|\beta-\pi|^2}{\beta}$ and second, its solution is non-unique. 

Finally, there are two possible KL constraints, forward KL $D_{KL}(\pi||\beta)\leq \delta$ and backward KL  $D_{KL}(\beta||\pi)\leq \delta$. The author in \cite{vuong2018supervised} have shown that the non-parametric solution for the backward constraint, which is implemented in TRPO, is $\pi(a|s)=\frac{\beta(a|s)\lambda(s)}{\lambda'(s)-A^{\beta}(s,a)}$ where $\lambda,\lambda'$ are chosen such that the KL constraint is binding. However, the backward KL does not restrict the ratio $\frac{|\beta-\pi|^2}{\beta}$ simply since actions with zero behavioral probabilities, i.e. $\beta_i=0$ do not contribute to the constraint (see below a simple example). On the other hand, the forward KL constraint which its non-parameteric solution is $\pi(a|s)=\frac{\beta(a|s)}{Z_{\lambda}(s)}e^{A^{\beta}(s,a)/\lambda(s)}$ indeed bounds the $\frac{|\beta-\pi|^2}{\beta}$ ratio since $\beta_i=0$ forces $\pi_i=0$. However, this bound depends on the specific size of the advantages and it differs from state to state, hence, it requires an advanced knowledge of the rewards or inferring $\lambda$ from the stochastic data. Moreover, the KL parameters $\delta$ or $\lambda$ are much less intuitive than strictly defining the upper and lower ratios $c_{\max},c_{\min}$. 

\subsubsection{Unbounded Probability Ratios of the TV and backward KL constraints}

To verify that TV and KL do not necessarily regulate the probability ratios, let's consider a tiny example of maximizing the objective function $J^{PPO}$ for a single state MDP with two actions $\{a_0,a_1\}$. Assume a behavior policy $\beta=\{1, 0\}$ and an estimated advantage $A^{\beta} = \{0, 1\}$. We search for a policy $\pi=\{1-\alpha, \alpha\}$ that Maximize improvement step under the TV or KL constraints.

For a $\delta$-TV constraint,  $\frac{1}{2}(1- (1-\alpha) + \alpha) = \alpha \leq \delta$. The improvement step in this case is $\sum_{a_i}A^{\beta}_i\pi_i = \alpha$. Hence the solution is $\alpha=\delta$ and the probability ratio $\pi(a_1) / \beta(a_1)$ is unconstrained (and undefined). Similarly for a backward $\delta$-KL constraint we get $-\log\alpha\leq\delta$ and the improvement step is identical. Hence $\alpha=e^{-\delta}$ and again, no constraint is posed on the probability ratio.

\subsection{Max-TV}
\begin{algorithm}[H]
\DontPrintSemicolon
\KwData{$s$, $\beta$, $A^{\mathcal{D}}$, $\delta$}
\KwResult{$\{\pi(a|s),\ a\in\mathcal{A}\}$}
\Begin{
$\pi(a|s) \longleftarrow \beta(a|s)$, $\forall a\in\mathcal{A}$\;
$a = \arg\max_{a\in\tilde{\mathcal{A}}} A^{\mathcal{D}}(s,a)$\;
$\Delta = \min\{\delta, 1-\beta(a|s)\}$\;
$\pi(a|s) \longleftarrow \pi(a|s) + \Delta$\;
$\mathcal{\tilde{A}} \longleftarrow \mathcal{A}$\;
\While{$\Delta > 0$}{
$a = \arg\min_{a\in\tilde{\mathcal{A}}} A^{\mathcal{D}}(s,a)$\;
$\Delta_a = \min\{\Delta, \beta(a|s)\}$\;
$\mathcal{\tilde{A}} \longleftarrow \mathcal{\tilde{A}}/a$\;
$\Delta \longleftarrow \Delta - \Delta_a$\;
$\pi(a|s) \longleftarrow \pi(a|s) - \Delta_a$\;
}
}
\caption{Max-TV}
\end{algorithm}

\subsubsection{The PPO Objective Function}

\citep{schulman2017proximal} suggested a surrogate objective function, termed \textit{Proximal Policy Optimization} (PPO), that heuristically should provide a reliable performance as in TRPO without the complexity of a TRPO implementation:
\begin{equation*}
J^{PPO}(\pi) = \mathbb{E}_{s\sim\beta}\left[ \sum_{a\in\mathcal{A}} \beta(a|s) \min\left( A^{\beta}(s,a)\frac{\pi(a|s)}{\beta(a|s)}, A^{\beta}(s,a)\clip\left(\frac{\pi(a|s)}{\beta(a|s)},1-\varepsilon,1+\varepsilon\right) \right) \right],
\end{equation*}
where $\varepsilon$ is a hyperparameter. PPO tries to ground the probability ratio by penalizing negative advantage actions with probability ratios above $1-\varepsilon$. In addition, it clips the objective for probability ratios above $1+\varepsilon$ so there is no incentive to move the probability ratio outside the interval $[1-\varepsilon,1+\varepsilon]$. However, we show in the following that the solution of PPO is not unique and is dependent on the initial conditions, parametric form and the specific optimization implementation. This was also experimentally found in \citep{henderson2017deep}. The effect of all of these factors on the search result is hard to design or predict. Moreover, some solutions may have unbounded probability ratios, in this sense, $J^{PPO}$ is not safe. 

First, notice that PPO maximization can be achieved by ad hoc maximizing each state since for each state the objective argument is independent and there are no additional constraints. Now, for state $s$, let's divide $\mathcal{A}$ into two sets: the set of positive advantage actions, denoted $\mathcal{A}^{+}$, and the set of negative advantage actions, $\mathcal{A}^{-}$. For convenience, denote $c_i = \frac{\pi(a_i|s)}{\beta(a_i|s)}$ and $\beta_i=\beta(a_i|s)$. Then, we can write the PPO objective of state $s$ as
\begin{equation*}
J^{PPO}(\pi,s) = \sum_{a_i\in\mathcal{A}^{+}} \beta_i A^{\beta}_i \min\left(c_i, 1+\varepsilon\right) - \sum_{a_i\in\mathcal{A}^{-}} \beta_i (-A^{\beta}_i) \max\left(c_i, 1-\varepsilon\right).
\end{equation*}
Clearly maximization is possible (yet, still not unique) when setting all $c_i=0$ for $a_i\in\mathcal{A}^{-}$, namely, discarding negative advantage actions. This translates into a reroute maximization with parameters $(c_{\min},c_{\max})=(0,1+\varepsilon)$ 
\begin{equation*}
\arg\max_{c_i}[J^{PPO}(\pi,s)] = \arg\max_{c_i}\left[\sum_{a_i\in\mathcal{A}^{+}} \beta_i A^{\beta}_i \min\left(c_i, 1+\varepsilon\right)\right] = \arg\max_{c_i}\left[\sum_{a_i\in\mathcal{A}^{+}} c_i \beta_i A^{\beta}_i \right]
\end{equation*}
for $c_i \leq 1+\varepsilon$. The only difference is that the sum $\sum_i c_i \beta_i=1-\Delta$ may be less then 1. In this case, let us take the unsafe approach and dispense $\Delta$ to the highest ranked advantage. It is clear that partition of $\Delta$ is not unique, and even negative advantage actions may receive part of it as long as their total probability is less than $(1-\varepsilon)\beta_i$. We summarize this procedure in the following algorithm.

\begin{algorithm}[H]
\label{alg:unsafe_ppo}
\DontPrintSemicolon
\KwData{$s$, $\beta$, $A^{\beta}$, $\varepsilon$}
\KwResult{$\{\pi(a|s),\ a\in\mathcal{A}\}$}
\Begin{
$\tilde{A}=\{\tilde{A}^{+}, \tilde{A}^{-}\}$\;
$\mathcal{\tilde{A}} \longleftarrow \mathcal{A}^{+}$\;
$\Delta \longleftarrow 1 $\;
$\pi(a|s) \longleftarrow 0 \ \forall a \in \mathcal{A}$\;
\While{$\Delta > 0$ and $|\mathcal{\tilde{A}}| > 0$} {
$a = \arg\max_{a\in\tilde{\mathcal{A}}} A^{\beta}(s,a)$\;
$\Delta_a = \min\{\Delta, (1+\varepsilon)\beta(a|s)\}$\;
$\mathcal{\tilde{A}} \longleftarrow \mathcal{\tilde{A}}/a$\;
$\Delta \longleftarrow \Delta - \Delta_a$\;
$\pi(a|s) \longleftarrow \pi(a|s) + \Delta_a$\;
}
$a = \arg\max_{a\in\mathcal{A}} A^{\beta}(s,a)$\;
$\pi(a|s) \longleftarrow \pi(a|s) + \Delta$\;
}
\caption{Ad hoc PPO Maximization}
\end{algorithm}

\section{Learning from Observations of Human players: Technical Details}
\label{method}
\subsection{Dataset Preprocessing}

The dataset \citep{kurin2017atari} (Version 2) does not contain an end-of-life signal. Therefore, we tracked the changes of the life icons in order to reconstruct the end-of-life signal. The only problem with this approach was in Qbert where the last life period has no apparent icon but we could not match a blank screen since, during the episode, life icons are flashed on and off. Thus, the last end-of-life signal in Qbert is missing. Further, the dataset contained some defective episodes in which the screen frames did not match the trajectories. We found them by comparing the last score in the trajectory file to the score that appears in the last or near last frame. 

We also found some discrepancies between the Javatari simulator used to collect the dataset and the Stella simulator used by the atari\_py package:
\begin{itemize}
\item
The Javatari reward signal has an offset of $-2$ frames. We corrected this shift in the preprocessing phase.
\item
The Javatari actions signal has an offset of $\sim-2$ frames (depending on the action type), where it is sometimes recorded non-deterministically s.t. the character executed an action in a frame before the action appeared in the trajectory file. We corrected this shift in the preprocessing phase. 
\item
The Javatari simulator is not deterministic, while Stella can be defined as deterministic. This has the effect that icons and characters move in different $mod_4$ order, which is crucial when learning with frame skipping of 4. Thus, we evaluated the best offset (in terms of score) for each game and sampled frames according to this offset.
\item
There is a minor difference in colors/hues and objects location between the two simulators. 

\end{itemize}

\subsection{Network Architecture}

A finite dataset introduces overfitting issues. To avoid overfitting, DQN constantly updates a replay buffer with new samples. In a finite dataset this is not possible, but, contrary to many Deep Learning tasks, partitioning into training and validation sets is also problematic: random partitions introduce a high correlation between training and testing, and blocking partitioning \citep{racine2000consistent} might miss capturing parts of the action-states space. Fortunately, we found that Dropout \citep{srivastava2014dropout}, as a source of regularization, improves the network's resiliency to overfitting. In addition, it has the benefit of better generalization to new unseen states since the network learns to classify based on a partial set of features. We added two sources of dropout: (1) in the input layer ($25\%$) and (2) on the latent features i.e. before the last layer ($50\%$).

We also found that for a finite dataset with a constant statistics, Batch Normalization (BN) can increase the learning rate. Therefore, we used BN layer before the ReLu nonlinearities. Note that for an iterative RL algorithm, BN may sometimes impede learning because the statistics may change during the learning process. To summarize, we used two DQN style networks: one for $\beta$ and the second for $Q^{\beta}$. We used the same network architecture as in \citep{mnih2015human} except for the following add-ons:

\begin{itemize}
\item
A batch normalization layer before nonlinearities.
\item
A 25\% Dropout in the input layer.
\item
A 50\% Dropout in the latent features layer. To compensate for the dropped features we expanded the latent layer size to 1024, instead of 512 as in the original DQN architecture.
\item
An additional output that represents the state's value.
\end{itemize}

\subsection{Hyperparameters tables}

\renewcommand{\arraystretch}{1.2}
\begin{table}[ht]
\centering
\caption{Policy and $Q$-value networks Hyperparameters}
\begin{tabular*}{8cm}[t]{lc}
\toprule
Name & Value\\
\midrule
Last linear layer size  & 1024\\
Optimizer & Adam\\
Learning Rate &  0.00025\\
Dropout {[}first layer{]} &  0.25\\
Dropout {[}last layer{]} &  0.5\\
Minibatch &  32\\
Iterations &  1562500\\
Frame skip &  4\\
Reward clip  &  -1,1\\
\bottomrule
\end{tabular*}
\end{table}

\renewcommand{\arraystretch}{1.2}
\begin{table}[ht]
\centering
\caption{DQfD Hyperparameters}
\begin{tabular*}{10cm}[t]{lc}
\toprule
Name & Value\\
\midrule
Optimizer & Adam\\
Learning Rate &  0.0000625\\
$n$-steps &  10\\
Target update period &  16384\\
$l_{a_E}$   &  0.8\\
Priority Replay exponent  &  0.4\\
Priority Replay IS exponent  &  0.6\\
Priority Replay constant  &  0.001\\
Other parameters  &  As in policy/value networks \\ & (except Dropout)\\
\bottomrule
\end{tabular*}
\end{table}

\subsection{Network Architecture}

A finite dataset introduces overfitting issues. To avoid overfitting, DQN constantly updates a replay buffer with new samples. In a finite dataset this is not possible, but, contrary to many Deep Learning tasks, partitioning into training and validation sets is also problematic: random partitions introduce a high correlation between training and testing, and blocking partitioning \citep{racine2000consistent} might miss capturing parts of the action-states space. Fortunately, we found that Dropout \citep{srivastava2014dropout}, as a source of regularization, improves the network's resiliency to overfitting. In addition, it has the benefit of better generalization to new unseen states since the network learns to classify based on a partial set of features. We added two sources of dropout: (1) in the input layer ($25\%$) and (2) on the latent features i.e. before the last layer ($50\%$).

We also found that for a finite dataset with a constant statistics, Batch Normalization (BN) can increase the learning rate. Therefore, we used BN layer before the ReLu nonlinearities. Note that for an iterative RL algorithm, BN may sometimes impede learning because the statistics may change during the learning process. To summarize, we used two DQN style networks: one for $\beta$ and the second for $Q^{\beta}$. We used the same network architecture as in \citep{mnih2015human} except for the following add-ons:

\begin{itemize}
\item
A batch normalization layer before nonlinearities.
\item
A 25\% Dropout in the input layer.
\item
A 50\% Dropout in the latent features layer. To compensate for the dropped features we expanded the latent layer size to 1024, instead of 512 as in the original DQN architecture.
\item
An additional output that represents the state's value.
\end{itemize}

The advantage for the PPO objective was estimated with $A_i = Q_i - \sum_i \beta_i Q_i$.

\subsection{Hyperparameters tables}

\renewcommand{\arraystretch}{1.2}
\begin{table}[ht]
\centering
\caption{Policy and $Q$-value networks Hyperparameters}
\begin{tabular*}{8cm}[t]{lc}
\toprule
Name & Value\\
\midrule
Last linear layer size  & 1024\\
Optimizer & Adam\\
Learning Rate &  0.00025\\
Dropout {[}first layer{]} &  0.25\\
Dropout {[}last layer{]} &  0.5\\
Minibatch &  32\\
Iterations &  1562500\\
Frame skip &  4\\
Reward clip  &  -1,1\\
\bottomrule
\end{tabular*}
\end{table}

\renewcommand{\arraystretch}{1.2}
\begin{table}[ht]
\centering
\caption{DQfD Hyperparameters}
\begin{tabular*}{10cm}[t]{lc}
\toprule
Name & Value\\
\midrule
Optimizer & Adam\\
Learning Rate &  0.0000625\\
$n$-steps &  10\\
Target update period &  16384\\
$l_{a_E}$   &  0.8\\
Priority Replay exponent  &  0.4\\
Priority Replay IS exponent  &  0.6\\
Priority Replay constant  &  0.001\\
Other parameters  &  As in policy/value networks \\ & (except Dropout)\\
\bottomrule
\end{tabular*}
\end{table}

\renewcommand{\arraystretch}{1.2}
\begin{table}[ht!]
\centering
\caption{Final scores table}
\begin{tabular*}{12cm}[t]{lrrrr}
\toprule
Method & MsPacman & Qbert & Revenge & SpaceInvaders \\
\midrule
 Humans               &   3024    & 3401    &   1436  &   634\\
 Behavioral cloning   &   1519    & 8562    &  1761  &   678\\
 Reroute-$(0.5,1.5)$  &   1550    & 12123   &  2379  &   \textbf{709}\\
 Reroute-$(0.25,1.75)$  &   \textbf{1638}    & 13254   &  \textbf{2431}  &   682\\
 Reroute-$(0,2)$  &    1565   & \textbf{14494}   &  473   &   675\\
 TV$(0.25)$           &   1352    & 5089    &  390   &   573\\
 PPO$(0.5)$           &   1528    & 14089   &  388   &   547\\
 DQfD                 &   83      & 1404    &  1315  &   402\\
\bottomrule
\end{tabular*}
\end{table}

\newpage

\section{RL with RBI}

The $Q$-net architecture is identical to Duel DQN \cite{wang2015dueling} and the $\pi$-net architecture is identical to the advantage branch in Duel DQN with an additional final softmax layer. $Q$-net and $\pi$-net are saved to ram-disk every 100 batches and each actor process loads a new snapshot every 1K environment steps (this is slightly different from Ape-X where policies are loaded from the learner every 400 environment frames). The target-$Q$ is updated every 2500 training batches (as in the original paper). The actors' samples are stored in the dataset only at episode termination and they are sampled uniformly by the learner ones every 5K batches. This design was chosen in order to facilitate the implementation and it clearly generates lags in the training process. Nevertheless, we found that those lags are not necessarily negative since they serve as a momentum for the learning process where past networks are used to generate samples and past samples are used in the learning process. Such learning momentum can prevent the network from converging to a local optimum, however, we leave the trade-off between fast updates and learning momentum for future research. Another parameter that may generate lags and momentum is the pace of samples stored in the dataset by the parallel actors. We tried to control this parameter to an average of 240-280 new states for a single training batch in the learner. Learning rate and optimization parameters are taken from \cite{hessel2017rainbow}, i.e. Adam optimization with a learning rate of $0.00025/4$ for both networks. We do not clip gradient norm.
 
For training, as in Ape-X, we also capped the episode length to 50K frames. This helps hard exploration games such as Berzerk but it limits the score of easier games such as Spaceinvaders, Enduro and Asterix. The latter games may be considered as solved, converged into near optimal policy which rarely loses life and its score is limited only by the allowed episode length. For final score evaluation we capped the episode to 108K frames as in Rainbow and started the environment with the no-op initialization method. In the table below we list the final score across the 12 environments. The column of Ape-X represents the Ape-X baseline implementation as described in the experimental setup. The Rainbow score is taken as a reference from \cite{hessel2017rainbow}. Surprisingly, our Ape-X baseline provided better results from the original Ape-X paper \cite{horgan2018distributed} in 5 games: Asterix, Enduro, Freeway, IceHockey and Kangaroo, in much less wall clock time (roughly 24 hours) and significantly lower computer resources (a single machine with 48 cores and 4 Nvidia 1080Ti GPUs). We hypothesize that the different learning momentum may be the staple reason for that.

\renewcommand{\arraystretch}{1.2}
\begin{table}[ht]
\centering
\caption{Hyperparameters table}
\begin{tabular*}{8cm}[t]{lc}
\toprule
Name & Value\\
\midrule
$c_{\max}$ & 2\\
$c_{\min}$ & 0.1\\
$c_{greedy}$ &  0.1\\
$c_{mix}$ &  $\frac{m}{43.7 + m}$\\
priority exponent $\alpha$ &  0.5\\
Batch & 128\\
$N$ players & 256\\
\bottomrule
\end{tabular*}
\end{table}%
Note that $m$ is the number of actions.

\begin{figure}[ht!]
  \centering
  \includegraphics[width=1\linewidth]{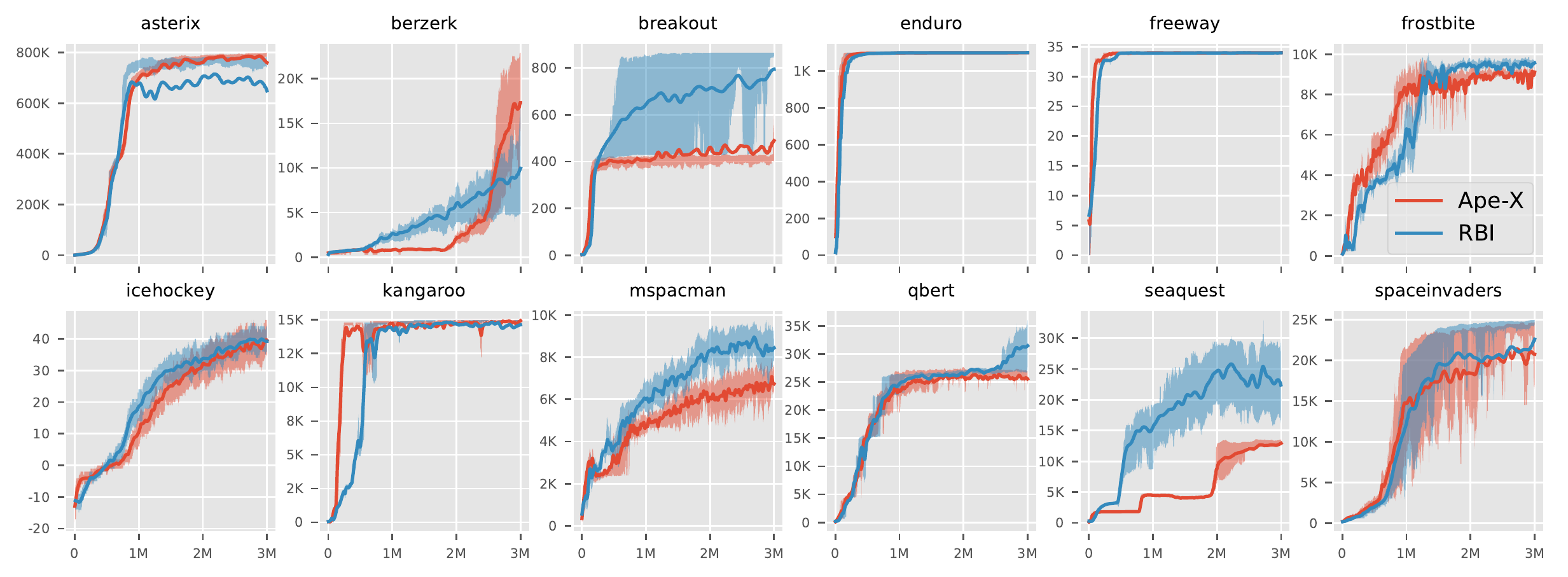}
  \caption{Performance curves of 12 Atari games. The second and third quartiles are shadowed.}
\end{figure}

\begin{table}[ht!]
\centering
\caption{Final scores table}
\label{table:rbi_score}
\begin{tabular*}{8cm}[t]{lrrr}
\toprule
Game & RBI & Ape-X & Rainbow \\
\midrule
Asterix & 838,018.3 & \textbf{926,429.2} & 428,200.3\\
Berzerk & 12,159.3 & \textbf{43,867.7} & 2545.6\\
Breakout & \textbf{780.4} & 515.2 & 417.5\\
Enduro & 2260.9 & \textbf{2276.1} & 2,125.9\\
Freeway & 34.0 & 34.0 & 34.0\\
Frostbite & \textbf{9,655.0} & 8,917.3 & 9,590.5\\
IceHockey & \textbf{40.4} & 39.9 & 1.1\\
Kangaroo & 14,426.7 & \textbf{14,897.5} & 14,637.5\\
MsPacman & \textbf{8,193.9} & 6,401.8 & 5,380.4\\
Qbert & 32,894.2 & 26,127.5 & \textbf{33,817.5}\\
Seaquest & \textbf{26,963.7} & 12,920.7 & 15,898.9\\
SpaceInvaders & \textbf{45,155.8} & 40,923.4 & 18,729.0\\
\bottomrule
\end{tabular*}
\end{table}%

\end{document}